\documentclass{bmvc2k}

\title{Recognising BSL Fingerspelling in Continuous Signing Sequences}

\addauthor{Alyssa Chan*}{}
{1}
\addauthor{Taein Kwon*}{}
{1}
\addauthor{Andrew Zisserman}{}
{1}

\addinstitution{
 Visual Geometry Group\\
 Department of Engineering Science\\
 University of Oxford\\
 UK\\
}

\usepackage{lineno} 

\usepackage{inconsolata}
\usepackage{bmvc2k_natbib}

\usepackage{graphicx}
\usepackage{hyperref}
\usepackage{booktabs}
\usepackage{tabularx}

\usepackage{pifont} 
\usepackage{amssymb}
\usepackage{tikz}
\usepackage{cuted}   
\usepackage[dvipsnames]{xcolor}
\usepackage{amsmath}
\usepackage{cleveref}

\definecolor{RoyalPurple}{RGB}{120,81,169}

\newcommand{\xmark}{\ding{55}}
\newcommand{\cmark}{\ding{51}} 
\newcommand{\dataset}{FS23K}

\runninghead{Chan*, Kwon*, Zisserman}
{FS23K - Recognising BSL Fingerspelling}
\usepackage{subcaption}
\begin{document}

\maketitle

\renewcommand\thefootnote{}
\footnotetext{* Equal contribution}
\begin{abstract}
Fingerspelling is a critical component of British Sign Language (BSL), used to spell proper names, technical terms, and words that lack established lexical signs. 
Fingerspelling recognition is challenging due to the rapid pace of signing and common letter omissions by native signers, while existing BSL fingerspelling datasets are either small in scale or temporally and letter-wise inaccurate.
In this work, we introduce a new large-scale BSL fingerspelling dataset, \dataset{}, containing 23K words and their intervals with letter-level annotations. The dataset is constructed automatically using a semi-supervised iterative annotation framework. In addition, we propose a fingerspelling recognition model that explicitly accounts for bi-manual interactions and mouthing cues. As a result, with refined annotations, 
our approach {\em halves} the character error rate (CER) on the CSLR dataset compared to the prior state of the art on fingerspelling recognition.
These findings demonstrate the effectiveness of our method and highlight its potential to support future research in sign language understanding and scalable, automated annotation pipelines. The project page can be found at \url{https://taeinkwon.com/projects/fs23k}.

\end{abstract}

\section{Introduction}

British Sign Language (BSL) is a visual-gestural language primarily used by the Deaf community in the United Kingdom \cite{sutton-spence_variation_1990}.
BSL possesses a distinct grammatical structure and syntax, independent of spoken English \cite{brennan_british_1979} and conveys meaning through a combination of lexical signs, gestures, facial expressions and fingerspelling. 
Lexical signs constitute the core vocabulary of the language and are produced through specific combinations of handshape, movement, location, and orientation.

Fingerspelling, which represents individual letters of
the alphabet through hand configurations, is a crucial key component of BSL.
Fingerspelling accounts for approximately 7\% of casual signing~\cite{raude_tale_2024},
and is commonly used to express proper names, place names, and technical terms that lack established lexical signs~\cite{sutton-spence_variation_1990}.
Native signers often employ abbreviations or reductions, for example shortening “Darwin” to “D” after its initial mention. 

\begin{figure}
    \centering
    \includegraphics[width=1.0\linewidth]{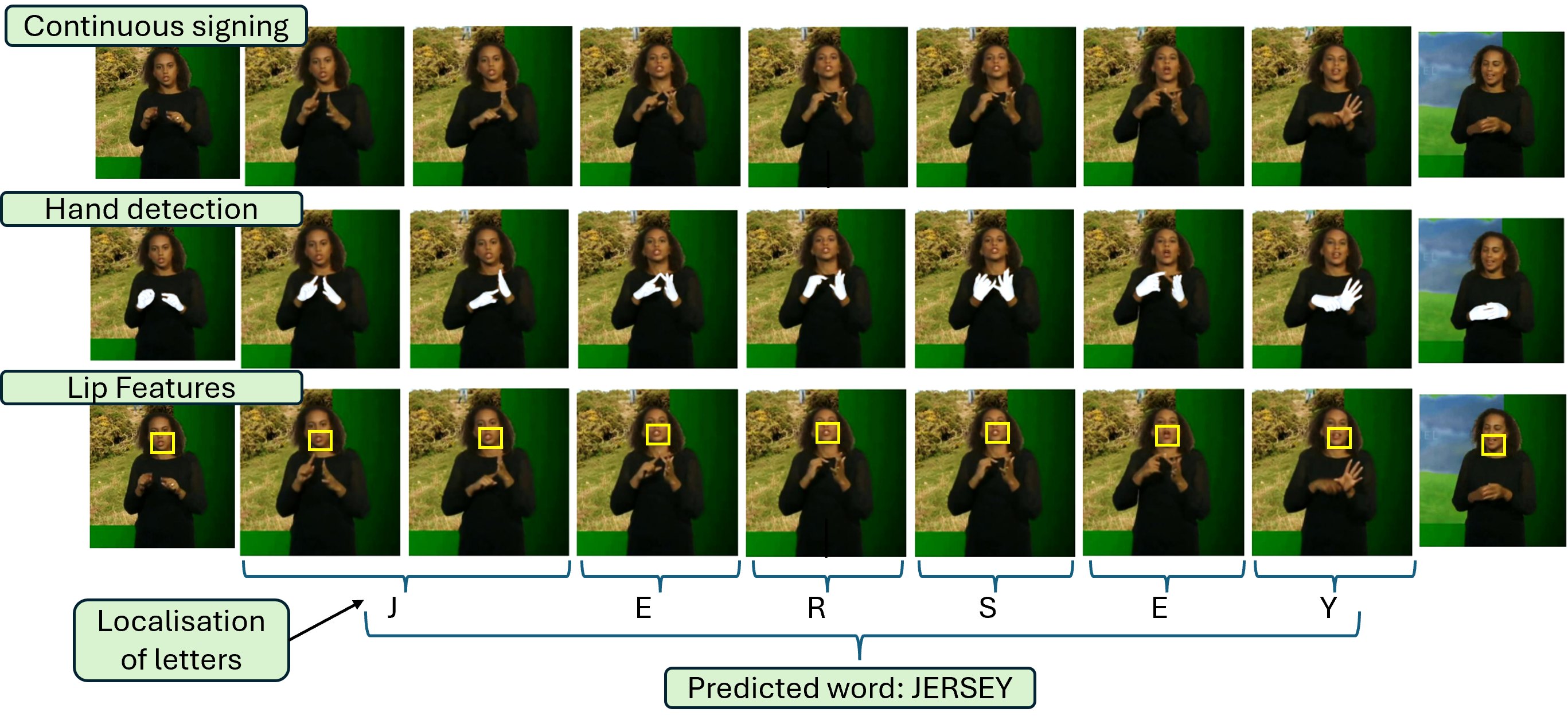}
    \vspace{0.02\textwidth}
    \caption{\small BSL fingerspelling recognition. Video frames from continuous signing where a fingerspelling temporal interval is detected, and hand and lip features are used to correctly recognize the signed letters.}
    \label{fig:teaser}
\end{figure}

Despite its importance, automatic BSL fingerspelling recognition from video as in Figure~\ref{fig:teaser}, which enables machines to understand and translate fingerspelling, remains highly challenging.
One major difficulty arises from the speed of articulation. 
Given a typical video frame rate of 25 frames per second, each letter is visible for only a small number of frames (approximately 2.5 frames on average), resulting in motion blur and limited visual evidence~\cite{wheatland_analysis_2016}. 

Fingerspelt letters may vary significantly due to co-articulation, and also vary across individuals in handshape, orientation, motion dynamics, and overall signing style, introducing substantial diversity that must be handled by robust recognition systems~\cite{sutton-spence_variation_1990}.

\begin{figure}[t]
    \centering
    \includegraphics[width=0.7\linewidth]{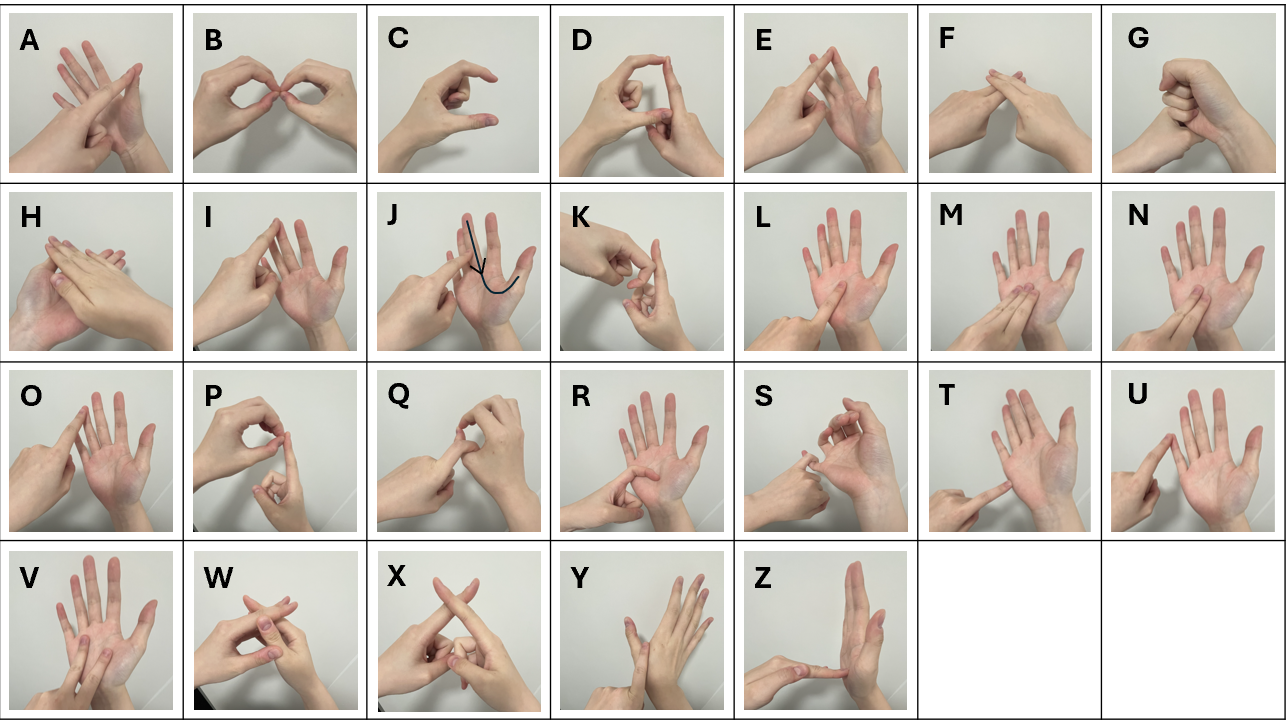}
    \vspace{2.5mm}
    \caption{The British Sign Language (BSL) fingerspelling alphabet. Unlike many other sign languages, BSL employs bi-manual fingerspelling, which poses additional challenges for recognition due to frequent occlusions between the two hands. Note, these examples are for a left-handed signer.
    }
    \vspace{0.01\textwidth}
    \label{fig:alphabet}
\end{figure}

To address these challenges, BSL-specific fingerspelling datasets are essential. While the largest existing fingerspelling datasets focus on American Sign Language (ASL)~\cite{shi2018american,shi2019fingerspelling,georg2025fsboard}, ASL differs fundamentally from BSL in both linguistic structure and articulation. Notably, BSL employs two-handed fingerspelling (see Figure~\ref{fig:alphabet}), which introduces additional challenges such as inter-hand occlusion and increased co-articulation~\cite{liwicki_automatic_2009}, compared to ASL where fingerspelling involves only a single hand. 
Furthermore, most existing datasets are restricted to isolated fingerspelling, failing to reflect the complexity of continuous, naturally occurring signing. 
Although a small number of BSL fingerspelling datasets have been proposed, they are either limited in scale \cite{raude_tale_2024} or rely solely on machine-generated annotations~\cite{prajwal2022weakly}, limiting their reliability for training and evaluation.

In this work, we address these limitations by introducing a large-scale, fine-grained annotation of continuous BSL fingerspelling, \dataset{}. 
Our annotations are derived from BBC broadcast content~\cite{albanie2021bbc}, ensuring coverage of in-the-wild, naturally occurring fingerspelling with realistic variability in signers, contexts, and visual conditions. 
We employ a multi-stage iterative annotation pipeline, leveraging hand keypoints extracted from RGB video using an off the shelf hand pose estimation model~\cite{pavlakos_reconstructing_2023}.
As a result, we obtain the largest annotations of continuous BSL fingerspelling to date, comprising 23K finely annotated fingerspelling instances with letter interval transition information and 154K fingerspelling words with temporal boundaries. 
Furthermore, as shown in Figure~\ref{fig:teaser}, we propose a Transformer-based fingerspelling recognition model that integrates hand keypoint and lip motion features using connectionist temporal classification (CTC) loss, achieving a character error rate (CER) of 0.250, representing an improvement of 0.331 CER over existing methods. 

In summary, our contributions are threefold: 
(i) we propose a Transformer-based fingerspelling recognition model that achieves state-of-the-art performance on BSL fingerspelling recognition on two public benchmarks; 
(ii) we introduce a multi-stage, iterative annotation pipeline that progressively refines noisy annotations to improve annotation quality;
(iii) we produce a large-scale BSL fingerspelling dataset, \dataset{}, comprising 154K fingerspelling instances with accurate temporal localization, along with 23K finely annotated fingerspelt words with letter level annotation. interval transition information. The trained models and dataset are publicly released.

\section{Related Work}

\subsection{Fingerspelling Datasets}

Several datasets have been proposed for fingerspelling recognition across different sign languages, as summarized in Table~\ref{tab:fingerspelling_datasets}. The most prominent fingerspelling datasets are available for American Sign Language (ASL). In particular, ChicagoFSWild~\cite{shi2018american} and ChicagoFSWild+~\cite{shi2019fingerspelling} are among the most widely used benchmarks. Their annotations are manually verified and sourced from YouTube videos. The diversity of video content improves robustness to real-world conditions. More recently, FSBoard~\cite{georg2025fsboard} was collected in a controlled environment by recruiting Deaf signers to record fingerspelling phrases using a selfie camera. Beyond ASL, fingerspelling datasets have also been released for other sign languages, including FGSL (Greek)~\cite{lata2024one}, One-Stage-TFS (Thai)~\cite{papadimitriou_multimodal_2024}, Auslan-Daily (Australian)~\cite{shen2023auslan}, and Bukva (Russian)~\cite{kapitanov2023slovo}. 

\begin{table*}[t]
 \caption{\small Fingerspelling datasets. 
 `Tem.\ boundary' refers to how tightly the fingerspelling segments are localised in time.
 `GT Letters' indicates whether the provided annotations correspond to the exact fingerspelt letters or to the associated word being fingerspelt.}
\vspace{0.02\textwidth}

\centering
\resizebox{\textwidth}{!}{%
\begin{tabular}{lccccc}
\toprule
\textbf{Dataset} & \textbf{Language} & \textbf{Anno.\ Method} & \textbf{Tem.\ Boundary} & \textbf{GT Letters} & \textbf{Scale}\\

\midrule

Auslan-Daily~\cite{shen2023auslan} 
& Auslan 
& Subtitle-aligned 
& Tight word-level 
& \xmark
& 2K clips \\
Bukva~\cite{kapitanov2023slovo} & RSL 
& Recording 
& Tight letter-level 
& \cmark
& 3.7K letters \\

One-Stage-TFS~\cite{lata2024one} &  TSL
& Recording 
& Img-level 
& \cmark
& 7.2k letters \\

\midrule
ChicagoFSWild~\cite{shi2018american} & ASL 
& Manual 
& Tight word-level 
& \cmark
& 38K letters   \\
ChicagoFSWild+~\cite{shi2019fingerspelling}   
& ASL 
& Manual 
& Tight word-level 
& \cmark
& 0.3M letters \\
FSboard~\cite{georg2025fsboard}   
& ASL 
& Recording
& Tight phrase-level
& \cmark
& 3.2M letters \\
\midrule

Transpeller~\cite{prajwal2022weakly} & BSL & Semi-manual & Loose word-level & \cmark & 5K words \\
CSLR~\cite{raude_tale_2024} & BSL & Manual & Near/close word-level & Partial & 3.4K words  \\

\dataset{} (ours) & BSL & Iterative Anno. & Tight word-level & \cmark & 23K words  \\
\bottomrule
\end{tabular}
} 
\label{tab:fingerspelling_datasets}
\end{table*}

\subsection{BSL Datasets}\label{sec:bsl}
In British Sign Language (BSL), which we focus on in this work, one of the first widely accessible datasets is the BBC-Oxford British Sign Language dataset (BOBSL)~\cite{albanie2021bbc}. This dataset contains over 1,400 hours of BSL-interpreted BBC broadcast footage, featuring 37 different signers and approximately 1.2 million sentences spanning a wide range of topics. All videos are recorded at 25 fps. However, BOBSL itself does not include explicit fingerspelling annotations. 

For fingerspelling annotations, using the BOBSL dataset, Transpeller~\cite{prajwal2022weakly} provides automatically generated, noisy training annotations (approximately 157K words) and semi-manually annotated test annotations (5K words). Semi-manual refers to the fingerspelt intervals being identified automatically, and the letter annotations being manually annotated. 
However, these annotations suffer from several limitations, including:  their limited scale of annotations, imprecise temporal localization, errors arising from reliance on mouthing cues, the absence of letter-level annotations, and failure cases in which multiple fingerspelling instances are incorrectly grouped together.
More recently, CSLR~\cite{raude_tale_2024} introduces a smaller set of  fingerspelling instances (1.9K words for training and 1.4K words for testing),  manually annotated by native BSL deaf signers, and with more accurate temporal boundaries, addressing some of the shortcomings of the Transpeller dataset. 
CSLR annotations only contain ground truth (GT) letters for a subset of the fingerspelt words.

This reduces the training and test set to 1.8K and 0.8K.

\subsection{Fingerspelling Detection and Recognition}
In American Sign Language (ASL), fingerspelling recognition has been actively studied. Early approaches relied on handcrafted features extracted from RGB videos~\cite{goh_dynamic_2006, liwicki_automatic_2009}, depth sensors~\cite{kisacanin_recognition_2005, rambhau_recognition_2013, kumar_deaf-bsl_2022}, and wearable sensors~\cite{oz_american_2011}. 
Hidden Markov Models (HMMs) were commonly used to model the temporal dynamics of fingerspelling gestures~\cite{goh_dynamic_2006, liwicki_automatic_2009, rambhau_recognition_2013, ricco_fingerspelling_2010}.
More recently, deep learning has significantly advanced fingerspelling recognition. Modern approaches detect and track hands and fingers~\cite{pinnington_machine_2024, papadimitriou_seeing_2025, low_hands-_2025}, and leverage convolutional neural networks (CNNs)~\cite{kumar_deaf-bsl_2022, papadimitriou_multimodal_2024}, recurrent neural networks (RNNs)~\cite{danko_recognition_2019}, and Transformer-based architectures~\cite{fayyazsanavi_fingerspelling_2024, prajwal_prajwalkrtranspeller_2025} to model spatiotemporal information.

Despite this progress, BSL fingerspelling recognition remains challenging due to the language-specific characteristics of the use of two hands, which limit the direct transferability of ASL-focused methods to BSL. 
As a result, BSL fingerspelling recognition in continuous signing sequences  has been comparatively underexplored. 
One of the few works addressing BSL fingerspelling in continuous signing is Transpeller~\cite{prajwal2022weakly}, which  proposes a weakly supervised, multi-stage training strategy. 
In the first stage, a small number of manually labeled exemplar frames are used to identify candidate fingerspelling segments. In the second stage, the model refines fingerspelling localization and letter predictions using automatically generated annotations from mouthings.
In contrast to this approach, our method iteratively constructs accurate fingerspelling annotations by leveraging lightweight 3D hand keypoints, explicit modeling of bi-manual relationships, and lip features. This design enables scalable and precise annotation.

\begin{figure*}
    \centering
    \includegraphics[width=0.9\textwidth]{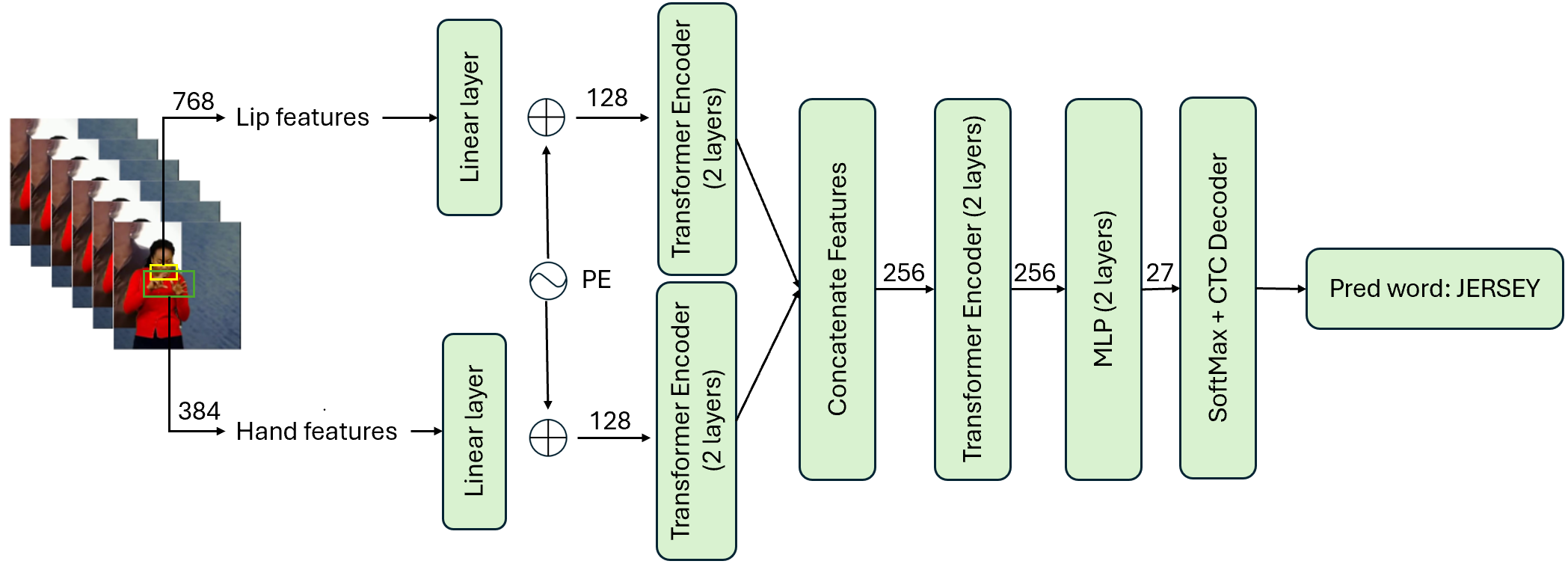}
    \vspace{0.01\textwidth}
    \caption{\small{Fingerspelling recognition network architecture. The model leverages two complementary feature modalities: lip features extracted using Auto-AVSR~\cite{ma2023auto} and hand features obtained from HaMeR~\cite{pavlakos_reconstructing_2023}. 
    Each modality is first passed through an individual linear projection to align feature dimensions, followed by separate Transformer encoders. 
    The encoded features are then concatenated and further processed by a Transformer encoder. 
    Finally, a two-layer MLP predicts per-frame letter labels, which are also used as inputs to the CTC decoder. The dimensions are shown in the numbers beside the arrows. The 384 dimensional hand feature vector represents both hands.}}
    \label{fig:architecture}
\end{figure*}

\section{Fingerspelling Recognition Model}\label{model}
Unlike many other sign languages, BSL fingerspelling uses both hands rather than a single hand. Fingerspelling is also complemented by mouthing,  as often the associated word being fingerspelt will be mouthed. We first describe the overall architecture that integrates information from both hands and lip motion in Section~\ref{archtecture}, followed by the loss functions that consider the static shape and transitions in fingerspelling, detailed in Section~\ref{loss}.
Finally, we provide the implementation details in Section~\ref{implementation_details}.

\subsection{Architecture}
\label{archtecture}
The architecture employs a Transformer-based network that jointly models lip features and bi-manual hand features for fingerspelling recognition. 
Each feature modality is first projected into a Transformer encoder, and the resulting representations are then concatenated. Additional encoder layers are applied to further learn a unified representation. Finally, MLP layers are used to classify the letters. The input is the entire fingerspelling sequence and the output is the predicted word spelt in the interval. The architecture is illustrated in Figure~\ref{fig:architecture}. 

\paragraph{Hand Features}

Hand keypoints provide a concise representation of hand shape and motion. We employ an off-the-shelf 3D hand keypoint estimator, HaMeR~\cite{pavlakos_reconstructing_2023}, to extract 3D keypoints for both the left and right hands. Our hand features consist of the 3D positioning of 21 joints per hand, along with the 2D and 3D distances between the hand centers to capture the spatial relationship between the two hands. These features are extracted per-frame. A full description of the features is given in the supplementary material.

\paragraph{Lip Features.}

We extract lip features using the ResNet backbone trained for lip-reading in  Auto-AVSR~\cite{ma2023auto}. Specifically, we crop the mouth region to precisely capture lip shape and then resize the cropped images before feeding them into the ResNet.

\paragraph{Network.} Each modality is first projected to a 128-dimensional embedding using a single linear layer. We use two encoder layers, with 2 heads, hidden dimension of 512 and dropout probability of 0.3. We apply normalisation before the MLP layers. The MLP consists of 2 linear layers with ReLU and dropout (0.3) in between. 

\subsection{Loss functions}
\label{loss}

We use both per-frame and sequence level losses. The per-frame loss,
$L_f$, uses the cross-entropy between the per-frame letter ground truth and prediction. This captures the per-frame shape of the hands and lips.  However, per-frame supervision can be noisy due to the difficulty of precise frame-level annotation, and it does not explicitly model transitions between letters. 

To address these limitations and to model the transitions of the hands and lips, we additionally incorporate the CTC sequence level loss, $L_{\mathrm{CTC}}$~\cite{graves_connectionist_nodate}, which is well suited to  sequence prediction tasks.

\begin{equation}
\mathcal{L}_{\mathrm{CTC}} = - \log p(\mathbf{y} \mid \mathbf{x}),
\end{equation}

where the conditional probability is defined as

\begin{equation}
p(\mathbf{y} \mid \mathbf{x}) = \sum_{\boldsymbol{\pi} \in \mathcal{B}^{-1}(\mathbf{y})}
\prod_{t=1}^{T} p(\pi_t \mid \mathbf{x}),
\end{equation}

and $\mathcal{B}$ denotes the CTC collapse function that removes repeated letters and blank symbols.

The CTC loss allows for blank frames during co-articulation, effectively handling transition frames in which the hands and lips are in motion.

Finally, the overall training objective is defined as:
\begin{equation}
\mathcal{L} = \lambda_{f} \cdot \mathcal{L}_{f} + \lambda_{CTC} \cdot \mathcal{L}_{\mathrm{CTC}},
\end{equation}
where $\lambda_{f}$ denotes the coefficient for the per-frame cross entropy loss, and $\lambda_{CTC}$ denotes the coefficient for the CTC loss.

\subsection{Training Implementation Details}
\label{implementation_details}
We use a stride of 1, and concatenate the hand features of
three neighboring frames together for the final feature of the central frame. This allows us to use the motion of the hands between frames to improve recognition. 
To mitigate class imbalance in letter frequency in training, we oversample the less frequent letters. We augment the hand keypoints using the method described in the appendix, and apply the same augmentation technique for lip feature extraction as in Auto-AVSR~\cite{ma2023auto}. The per-frame loss coefficient $\lambda_{f}$ is set to 0.02 and the CTC loss coefficient $\lambda_{CTC}$ is set to 0.98. The model was trained using the Adam optimizer~\cite{adam2014method}  for 1000 epochs using a learning rate of \(1e^{-4}\) and a dropout rate of 0.3. Training took approximately 5 hours using a NVIDIA GeForce RTX 2080 Ti GPU.

\paragraph{Letter-level masking.}
During training, we employ a letter-level masking strategy in the CTC loss, so that only letters present in the associated word are considered. For example, if the associated word is `cat', the loss is computed only for the letters `c',`a',`t'. This encourages the model to focus on distinguishing letters within the target word, rather than unnecessarily penalizing predictions for letters that do not appear.

\paragraph{Data augmentation.}
We applied data augmentation to both the hand and lip features by adding Gaussian noise with a standard deviation of 0.005. Since 99.7\% of values in a Gaussian distribution fall within 3 standard deviations, this corresponds to approximately 0.0015, or around 12 pixels. Additionally, we incorporated temporal augmentations, include frame neighbour swapping with a probability of 0.05, dropout of 0.02, and frame duplication with a probability of 0.05. A visualization of this augmentation is provided in the supplementary material.

\section{Dataset}

In this section we first describe what the dataset consists of, and then describe its construction and verification. 

\subsection{Dataset Statistics}

\begin{table}[h]
  \centering
    \caption{\small Dataset statistics. \dataset{} provides large-scale annotations, covering a wide range of samples and diverse signers. }
    \vspace{0.02\textwidth}

  \begin{tabularx}{0.8\columnwidth}{c | c c c c c}
    \toprule
    \#tem.-boundary& \#words & \#letters & \#frames & hours & \#signers \\
     133{,}909& 23{,}073 &  139{,}031 & 905{,}744& 10.06 & 27 \\

    \bottomrule
  \end{tabularx}

  \label{tab:stats}
\end{table}

\begin{figure}[h]
    \centering

    \begin{minipage}{0.48\columnwidth}
        \centering
        \includegraphics[width=\linewidth]{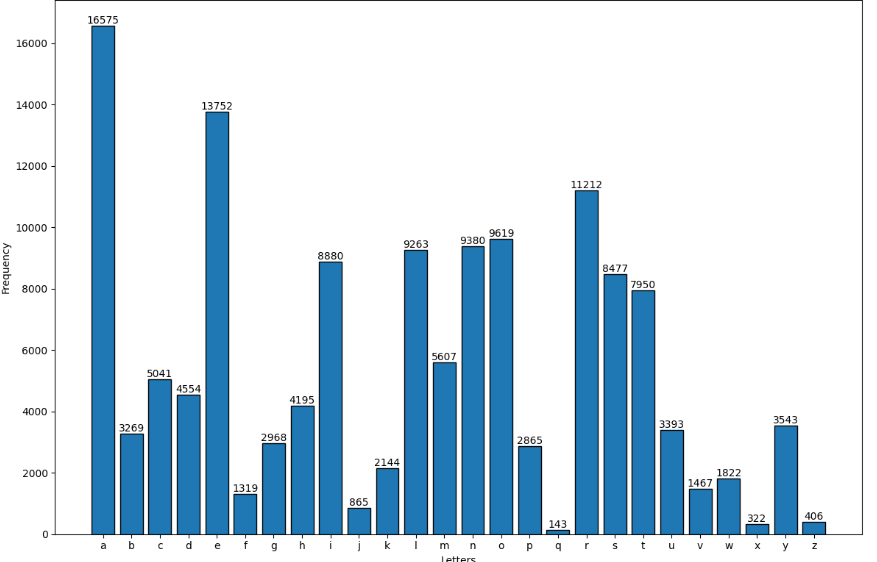}
        \vspace{0.00001\textwidth}
        \caption{\small Histogram of letter distribution in \dataset{}. The letters a (16,575) and e (13,752) occur most frequently, whereas q (143) and x (322) appear least often. This imbalance reflects the natural distribution of letters in in-the-wild BBC broadcast data, and similar distributions are also found in CSLR datasets.}
        \label{fig:histogram}
    \end{minipage}
    \hfill
    \begin{minipage}{0.48\columnwidth}
        \centering
        \includegraphics[width=\linewidth]{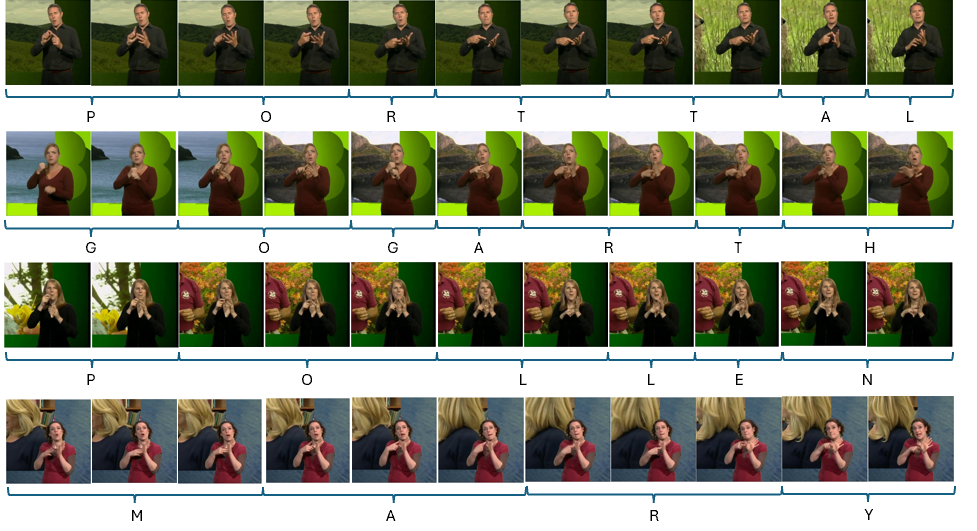}
        \vspace{0.0005\textwidth}
        \caption{\small Dataset samples. Thanks to our frame-wise and word-level classifier, \dataset{} provides precise temporal localization for each fingerspelt letter. This dataset contains diverse examples from the BOBSL dataset and covers a wide range of topics. }
        \label{fig:samples}
    \end{minipage}

\end{figure}

\dataset{} provides more precise and larger-scale annotations than existing BSL fingerspelling datasets.
Specifically, as shown in Table~\ref{tab:stats}, \dataset{} contains 23K fine-word fingerspelling instances with letter-level annotations and 133K fingerspelling temporal boundaries. 

We provide 133K tight temporal intervals where fingerspelling occurs, 66K of which have a corresponding word, although not all letters of this word will be present in the interval. For 23K of these intervals we provide the letters being fingerspelt and frame level annotation of these letters. Overall, \dataset{} comprises 23K words, 139K letter annotations, 920K frames, 10.5 hours of video, and 27 signers, supporting both the scale and diversity of the dataset.
Figure~\ref{fig:histogram} shows the distribution of letters in \dataset{}, which closely follows the natural letter frequency observed in BBC broadcast data from BOBSL.
Figure~\ref{fig:samples} illustrates representative samples, demonstrating the fine-grained letter-level annotations and precise per-frame temporal localization. 

\subsection{Dataset Construction}

Existing BSL fingerspelling training datasets suffer from limitations such as containing only a small number of high quality samples (the CSLR dataset), or loose annotations where the associated word is known but not the exact letters being spelt, and temporally inaccurate boundaries (the Transpeller dataset). 
On the other hand, in order to construct a new dataset, manual annotation of fingerspelling is both non-trivial and expensive, as it requires expert knowledge of BSL fingerspelling and precise temporal alignment. Moreover, simple alignment strategies that evenly divide frames among letters are insufficient, as letters vary in duration and abbreviations are frequently used in fingerspelling, so not all the letters of the word are spelt out. 

\begin{figure}
    \centering

    \begin{minipage}{0.48\columnwidth}
        \centering
        \includegraphics[width=\linewidth]{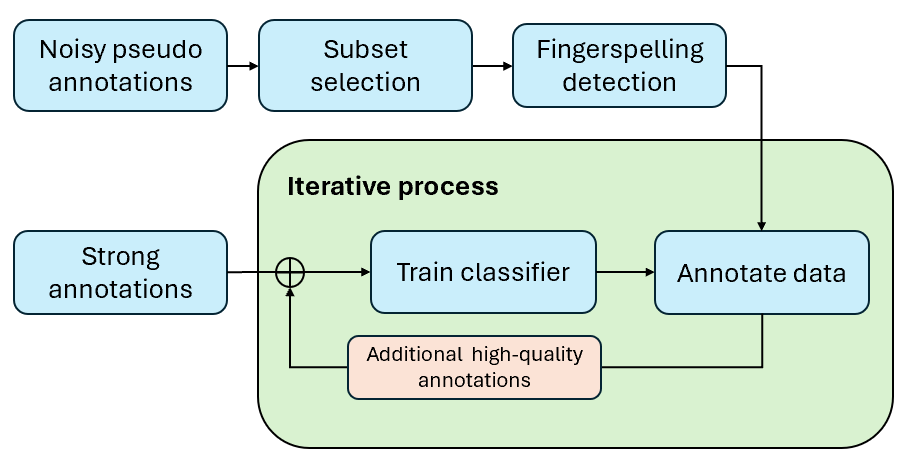}
        \vspace{0.005\textwidth}
        \caption{\small Dataset Construction. To construct \dataset{}, we employ a multi-stage iterative annotation strategy. Starting from a small set of strong annotations from the CSLR fingerspelling training set, we progressively re-annotate noisy pseudo annotations from the Transpeller dataset. During this iterative process, increasingly reliable annotations are accepted as the classifier improves through training on the expanding dataset.}
        \label{fig:iterative}
    \end{minipage}
    \hfill
    \begin{minipage}{0.48\columnwidth}
        \centering
        \includegraphics[width=\linewidth]{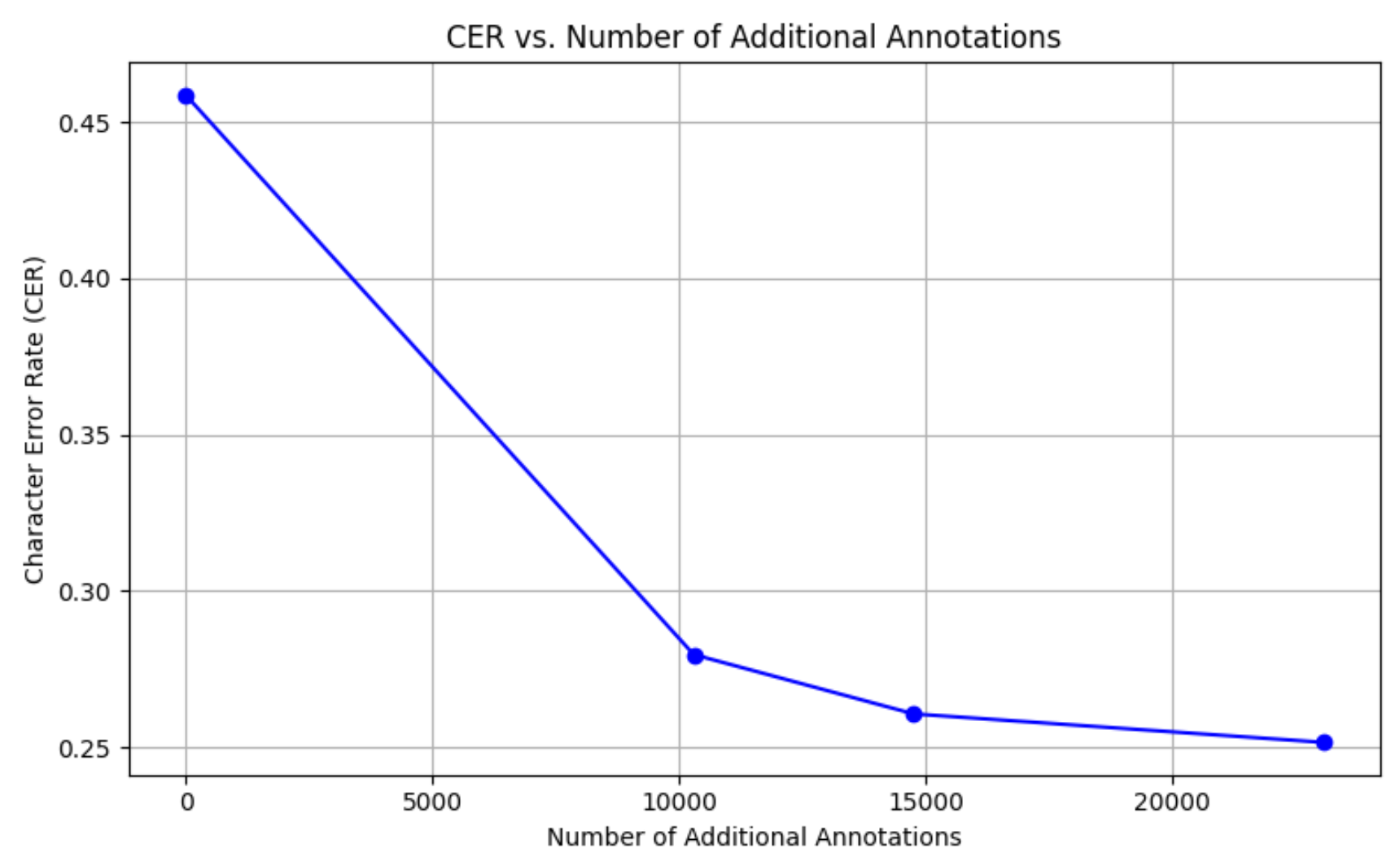}
        \vspace{0.05\textwidth}
        \caption{\small CER of CSLR test set vs.\ number of additional annotations from \dataset{}. This plot demonstrates the clear benefit of incorporating additional annotations, showing consistent improvements from the initial small set to the full set of refined annotations of \dataset{}.}
        \label{fig:cer_add_data}
    \end{minipage}

\end{figure}

Our objective is to construct a high quality dataset from the noisy annotations provided in the Transpeller dataset.
To address the above challenges, we adopt an {\em iterative re-annotation strategy}. Starting from a model trained on a small set of manually annotated samples, we progressively expand the number of high-quality annotations by using the current model to provide reliable annotations, and then re-training on these annotations. The process is illustrated in Figure~\ref{fig:iterative}.

Specifically, we initialize training with 1.8K strongly annotated words from the CSLR training dataset. Leveraging these annotations, we iteratively re-train per-frame and word-level classifiers to denoise the 157K large-scale pseudo-labeled word annotations from the Transpeller dataset. As a result, we annotate 23K high-quality fingerspelt words with letter interval transitions (Figure~\ref{fig:samples}). We detail the steps in the following. Note, our aim is to denoise the existing Transpeller annotations, not to detect missed finger spelling annotations.

\subsubsection{Strong Annotations}
We start from strong annotations provided by the CSLR dataset, where the exact letter sequence for each word is known. From the CSLR training set, we select 1.8K words (out of 1.9K) that include explicit letter annotations of the words. While CSLR does not provide precise frame-level letter annotations, we initialize letter interval transitions using an equal-number-of-frames-per-letter strategy (see Supplementary). This heuristic does not guarantee accurate per-frame alignment, but it provides a reasonable initialization for the subsequent iterative process.

\subsubsection{Annotation Source}
The Transpeller dataset is employed as a source of noisy pseudo annotations. From this set, we first extract a reliable subset by applying a series of filtering rules described in the supplementary material. However, since the Transpeller dataset provides automatically generated annotations, they often suffer from imprecise temporal localization of fingerspelling segments. 
In particular, the Transpeller dataset sometimes annotates segments where no fingerspelling is present, and incorrectly identifies two separate fingerspelling entries as one long interval.

\paragraph{Detection of Fingerspelling Intervals.}
\label{sec:detection}
We train a temporal fingerspelling detector for frame-level binary classification using the strong annotations from the CSLR dataset described above. We apply this detector to the videos with noisy pseudo Transpeller annotations to obtain precise fingerspelling intervals.  Detection examples are given in the supplementary material.

\paragraph{Detector implementation.}
For fingerspelling detection, we use the same hand feature vector described in Section~\ref{archtecture}. The detector is a 3-layer MLP binary classifier. Each layer consists of a linear layer (with 128, 64 and 32 units respectively), followed by layer normalization, ReLU and dropout. The final output is passed through a sigmoid function to classify the presence or absence of fingerspelling for each frame. We apply a 6-frame sliding-window with a stride of 1 frame. If at least 4 frames within this window are predicted as fingerspelling, we label the entire window as containing fingerspelling. 

\subsubsection{Iterative Process}
After refining the Transpeller fingerspelling intervals, we now have temporal intervals with a word associated with each, but we do not know which of the letters of the word are fingerspelt, or the frames corresponding to each letter. In the iterative process, we first train a frame-wise letter classifier using the strong annotations. We then use the frame-wise classifier to annotate the noisy psuedo annotations and iteratively improve the classifier. The letter predictions are aggregated for each fingerspelling interval to form a predicted word. Based on these refined word-level annotations, we further train a word-level classifier to obtain additional high-quality annotations. This procedure is repeated iteratively to progressively expand and refine the annotation set.

\paragraph{Frame-wise Classifier.}
To obtain per-frame letter annotations, we train a multiclass classifier with the same architecture as the detector described above (Section~\ref{sec:detection}), replacing the binary output with a multi-class prediction over letters. This classifier is then used to re-annotate the Transpeller samples by assigning letters based on a maximum-probability criterion.
Specifically, the classifier outputs a softmax probability distribution over the 26 letters (a-z) for each frame.
To avoid unreliable predictions, we only consider predictions where the maximum probability of the softmax exceeds a threshold of 0.35, the frame is then assigned the letter with highest probability. The threshold of 0.35 is determined heuristically, selecting the value that yields the most accurate and consistent predictions. Additional details on the maximum-probability criterion and alternative frame-acceptance strategies are provided in the supplementary material.
During re-annotation, we constrain the predicted letters to those appearing in the  word associated with the fingerspelling interval, as fingerspelt abbreviations correspond to a subset of the letters in the target word.

\paragraph{Word-level Classifier.}

Once sufficient confidence is achieved in the letter-level annotations, we train a word-level classifier using the same architecture as the main model described in Section~\ref{archtecture}, augmented with a CTC loss. Word-level refers to the exact letters spelt in the fingerspelling interval (which may not include all the letters in the word).
This stage requires full-letter annotations for each fingerspelling segment rather than frame-wise letter labels. As with the frame-level letter classifier, we perform iterative re-annotation and training. 

\paragraph{Iterative implementation.}

The process iterates three times, and at each iteration, newly accepted high-quality annotations are added to the training set. 
To construct word-level labels, we collapse consecutive repeated character predictions and blanks and concatenate the resulting sequence to form a candidate word. Double letters will have blanks in between so 'l-o-o-s-e' will return 'loose', - representing the blanks. We then compute the Character Error Rate (CER) between this predicted word and the corresponding associated word from the noisy pseudo annotations, retaining only entries with a CER below 0.3. 
Figure~\ref{fig:cer_add_data} illustrates the reduction in 
the CER as additional annotations are incorporated.
With only the CSLR training dataset, the model achieves a CER of 0.4472. 
As more data are progressively incorporated, the CER consistently decreases, demonstrating the effectiveness of our iterative re-annotation strategy.

\subsection{Verification}\label{verification}

To verify the quality of our generated annotations, human annotators manually and carefully evaluated the finger spelling for a subset of words in the FS23K dataset. 500 words were randomly sampled (approximately 2\% of the dataset) and the CER computed between the generated letter annotations and the manual (ground truth) letter annotations,  resulting in a CER of 0.0449.
We note that BSL fingerspelling is particularly challenging even for human annotators due to rapid signing, frequent hand occlusions, and incomplete articulation of letters, further underscoring the quality of our annotations.
\section{Results}

\subsection{Evaluation Datasets, Metrics, and Baselines}
\label{metrics}

\paragraph{Datasets.} We use the publicly available test sets from the Transpeller dataset~\cite{prajwal2022weakly} and the CSLR dataset~\cite{raude_tale_2024}. These are described in Section~\ref{sec:bsl}. Since the CSLR dataset contains partial ground-truth annotations, we carefully select a reliable subset of 0.8K samples to enable a precise comparison. The subset contains the exact letters present in the interval rather than only the associated word being fingerspelt. Note, the CSLR dataset keeps the BOBSL train/test divisions, so the test partition only has signers not seen during training.

\paragraph{Evaluation metrics.} Two metrics are used: {\it Character Error Rate (CER)} is calculated by counting the number of insertions \(I\), deletions \(D\), and substitutions \(S\) needed to transform the predicted letter sequence into the ground truth sequence (of character length \(N\)). 
This is calculated as ${\rm CER} = (S + D + I)/N $. A lower CER indicates a more accurate prediction, with 0 being perfect and 1 meaning no correct letters; and {\it Average Class Accuracy} is computed by averaging the per-class accuracy across all letters, reflecting letter-wise sensitivity. For each letter, we count the number of occurrences and compute the accuracy as the percentage of correctly identified instances.

\paragraph{Baselines.}
We use the publicly available Transpeller model~\cite{prajwal_prajwalkrtranspeller_2025} with a frame stride 
of~1. We also fine-tune this model on the FS23K dataset.
Note that we cannot include baselines from other sign languages, such as ASL, due to substantial differences in characteristics, including the use of two hands.

\subsection{Comparison on public testing datasets}

Results are given in Table~\ref{tab:cslrperf}. Our model achieves a CER of 0.250 and 0.389 on CSLR and Transpeller test sets, respectively.
There is a CER gap between the two test sets as the Transpeller dataset is noisy and contains long intervals ($\ge$ 3s) with multiple disjoint instances of fingerspelling. 
Since this is a public benchmark, we assess on it without correcting it for fair comparison with others. 
Our model significantly outperforms the Transpeller baseline by 0.331 on the CSLR test set and 0.144 on the Transpeller test set. 

\begin{table}[h]
\small
    \centering
        \caption{\small
        Comparison between Transpeller and our method. Our approach significantly improves performance on the CSLR and Transpeller test sets, achieving a reduction of 0.331 and 0.144 respectively in character error rate (CER).}
        \vspace{0.02\textwidth}
    \begin{tabular}{|c|c|c|}
    \hline
    \textbf{Methods}    &  \textbf{CER on CSLR test set} \(\downarrow\) & \textbf{CER on Transpeller test set} \(\downarrow\)  \\
    \hline
    Transpeller \cite{prajwal2022weakly} & 0.581 & 0.533\\
    Transpeller \cite{prajwal2022weakly} fine-tuned on \dataset{} & 0.549 & 0.496 \\
    Ours & \textbf{0.250} & \textbf{0.389}\\
    \hline
    \end{tabular}
\label{tab:cslrperf}
\end{table}

Fine-tuning the Transpeller baseline on the FS23K dataset decreases the CER from
0.581 to 0.549 on the CSLR test set and from 0.533 to 0.496
on the Transpeller test set. Since Transpeller was originally pretrained on 157K entries, fine-tuning it on the much smaller FS23K dataset (23K entries) naturally results in a more modest improvement.
Nevertheless, our method still outperforms this baseline, indicating the benefits brought in our model with fine-grained supervision and multimodal design.

\begin{figure*}[t]
\small
\centering

\begin{minipage}[t]{0.44\textwidth}
    \centering
    \includegraphics[width=0.9\linewidth]{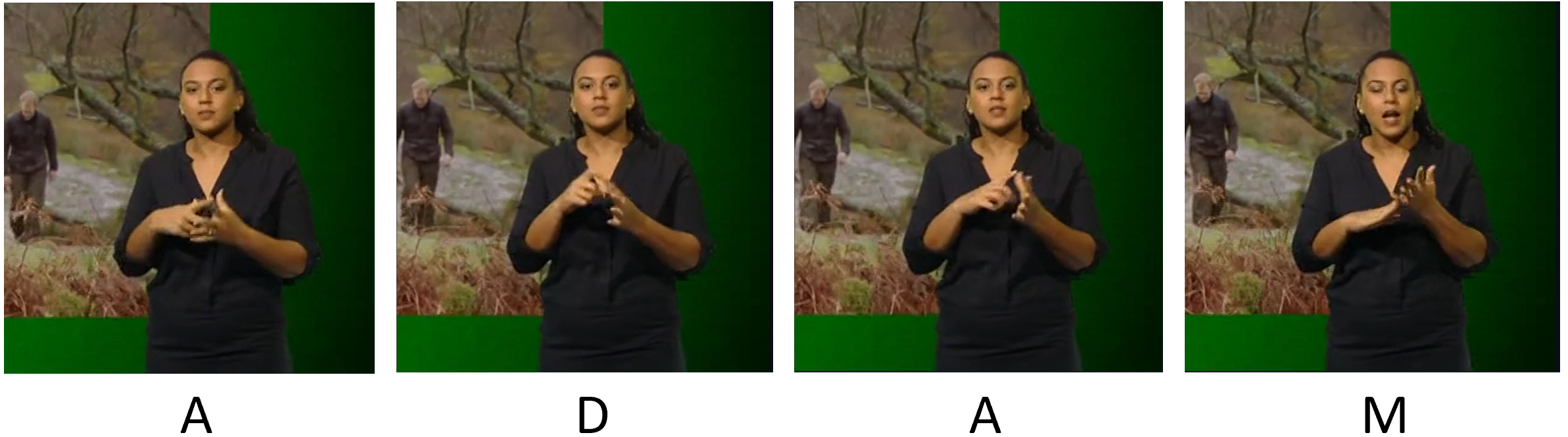}

    \begin{tabularx}{0.9\linewidth}{|X|X|c|}
    \hline
    \textbf{Source} & \textbf{Prediction} & \textbf{CER} \\
    \hline
    GT & adam & -- \\
    Transpeller &
    \textcolor{green}{a}\textcolor{blue}{--\,--}\textcolor{green}{m}
    & 0.5 \\
    Ours &
    \textcolor{green}{adam}
    & 0 \\
    \hline
    \end{tabularx}

    \vspace{4pt}
    \text{(a) Prediction for `adam'}
\end{minipage}
\hfill
\begin{minipage}[t]{0.52\textwidth}
    \centering
    \includegraphics[width=0.9\linewidth]{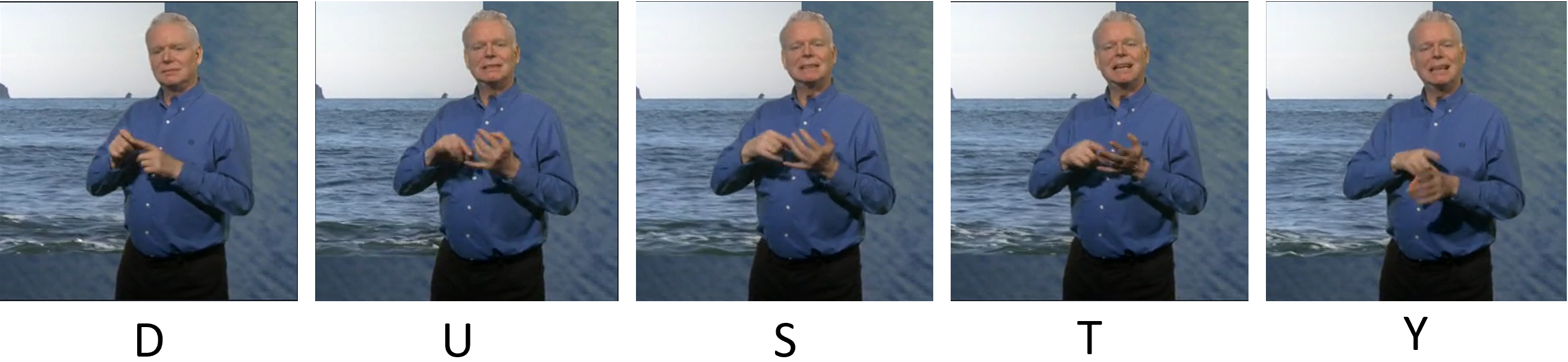}

    \begin{tabularx}{0.9\linewidth}{|X|X|c|}
    \hline
    \textbf{Source} & \textbf{Prediction} & \textbf{CER} \\
    \hline
    GT & dusty & -- \\
    Transpeller &
    \textcolor{red}{s}\textcolor{red}{a}\textcolor{green}{s}
    & 0.6 \\
    Ours &
    \textcolor{green}{dusty}
    & 0 \\
    \hline
    \end{tabularx}

    \vspace{4pt}
    \text{(b) Prediction for `dusty'}
\end{minipage}

\vspace{5pt}

\begin{minipage}[t]{1\textwidth}
    \centering
    \includegraphics[width=0.9\linewidth]{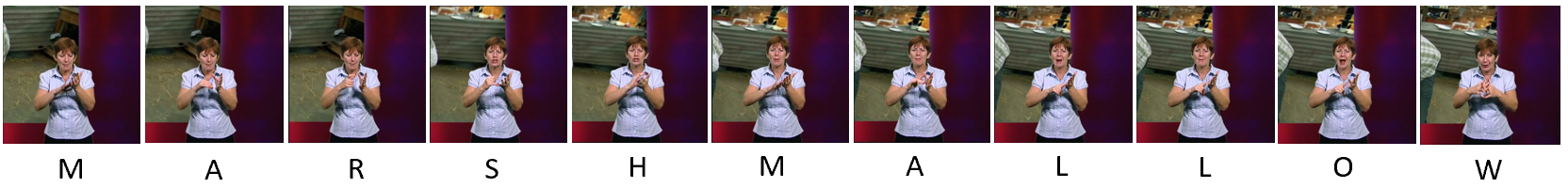}

    \begin{tabularx}{0.9\linewidth}{|X|X|c|}
    \hline
    \textbf{Source} & \textbf{Prediction} & \textbf{CER} \\
    \hline
    GT & marshmallow & -- \\
    Ours &
    \textcolor{green}{marshmallow}
    & 0 \\
    \hline
    \end{tabularx}

    \vspace{4pt}
    \text{(c) Prediction for `marshmallow'}
\end{minipage}

\vspace{6pt}

\caption{
Example correct fingerspelling predictions.
Top: a subset of the video frames with corresponding letters.
Bottom: full-word predictions from Transpeller and our model with character error rates (CER).
Colours indicate correct letters (\textcolor{green}{green}), substitutions (\textcolor{red}{red}), and insertions/deletions (\textcolor{blue}{blue}).
}
\label{fig:example_predictions}

\end{figure*}

In figure~\ref{fig:example_predictions}, successful examples are provided, showing a large improvement from the Transpeller model. In the first example, the final frame depicts the letter `m' with the fingers extended but not yet in contact with the palm. 
This example highlights the importance of modelling transitional frames, as hand motion and posture can already signal an `m' before full contact, which becomes harder to distinguish once the fingers touch the palm due to occlusion.

\begin{figure}
    \centering
    \includegraphics[width=0.9\linewidth]{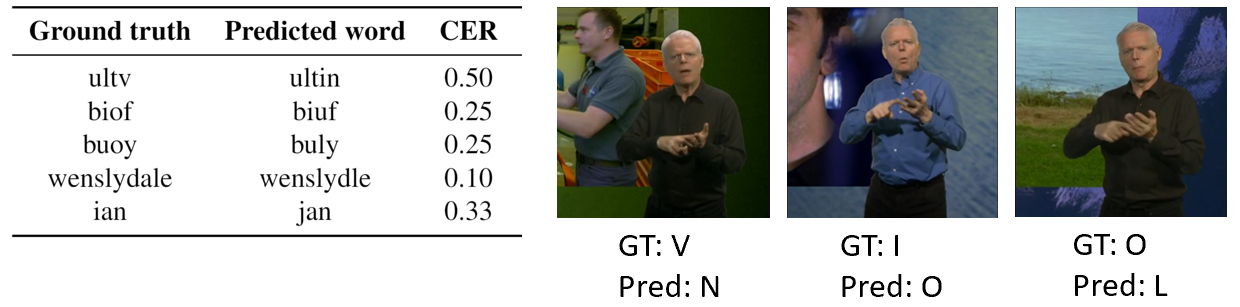}
       \vspace{0.02\textwidth} 
    \caption{Examples of ground truth and model predictions highlighting where our model struggles. The figures on the right are examples where letters can be easily confused with each other due to occlusion or similar hand positions. As previously noted, `n' and `v' are visually similar. Many letters involve the index finger outstretched and touching the other hand -- `l, t, a, e, i, o, u', and therefore these tend to have lower classification accuracies. This is shown in examples two and three.}
    \label{fig:badexamples}
\end{figure}

Figure~\ref{fig:badexamples} illustrates challenging examples.
In the first example (ultv), the model struggles with co-articulation and occlusions. During the transition from `t' to `v' the hand passes through intermediate letters resembling `i', which is occluded. In the second example (biof), the errors are due to vowel misclassifications, which is common due to occlusions and visual similarity. Example three (buoy) and four (wenslydle) show missed letters, due to the fast speed of signing, the exact letter position may not be fully formed and are most likely occluded. The letter `j' is a dynamic sign, meaning the letter is conveyed through motion and often moves through static letter positions, this is seen in example five.

\begin{figure}
    \centering
    \includegraphics[width=0.8\linewidth]{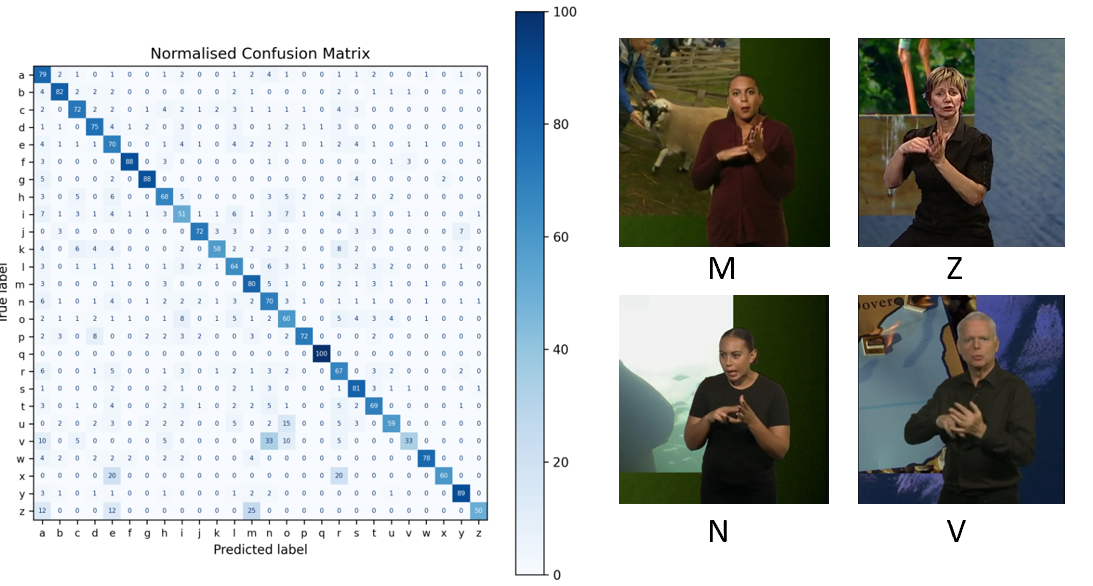}
    \vspace{6pt}
    \caption{Confusion Matrix: The darker blue areas indicate higher accuracy for each letter class, with the diagonals representing the percentage of true positives. All the off diagonal entries are false positives. The frames on the right show the letters that are often confused. The letters ’n’ and ’v’, differ in the configuration of two fingers on the palm – the fingers are together for ’n’ and apart for ’v’. This distinction is subtle and often obscured by occlusions. The letters 'm' and 'z' share an outstretched palm.}
    \label{fig:conf}
\end{figure}
Figure~\ref{fig:conf} illustrates the confusion matrix. Notably, the letters `i', `k', `l', `o', `v' and `z' exhibit lower performance than others due to their visual similarity. For example, `i' and `o' differ only in whether the index finger touches the middle finger or the ring finger of the other hand. Similar ambiguities also arise among `n', `v', and `z', `m', making them more prone to confusion. The most commonly signed vowels, `a' and `e', are predicted the most accurately as the model is more confident in these letters due to more training data.

\subsection{Ablation Studies}

\begin{table}[h]
\centering
\small

\caption{\small Ablation study. `Large-scale Anno.' indicates whether the model is trained on \dataset{}, whereas the absence of 'Large-scale Anno.' means that the model is trained only on the CSLR training dataset.
This ablation study shows that our model consistently improves both CER and average class accuracy on the CSLR test set, validating the effectiveness of our design choices.} 
\vspace{0.02\textwidth}

\resizebox{\columnwidth}{!}{%
\begin{tabular}{|c|c|c|c|c|c|}
\hline
\textbf{Large-scale Anno.} &  \textbf{Per-frame Loss} & \textbf{Weighted Sampling} & \textbf{Data Aug.} & \textbf{Average Class Acc.} & \textbf{CER$\downarrow$} \\
\hline
\xmark & \xmark & \xmark & \xmark & 46.92\% & 0.4472 \\
\xmark & \xmark & \cmark & \xmark & 50.48\% & 0.4490 \\
\xmark & \xmark & \cmark & \cmark & 52.61\% & 0.4447 \\
\xmark & \cmark & \cmark & \cmark & 53.03\% & 0.4396 \\
\cmark & \cmark & \cmark & \cmark & 70.65\% & 0.2497 \\
\hline
\end{tabular}%
}

\label{tab:ablation}
\end{table}

Table~\ref{tab:ablation} presents an ablation study examining the impact of incorporating the additional \dataset{} annotations, combining CTC and cross-entropy (CE) losses, and applying oversampling and data augmentation. The baseline model, trained without these enhancements, achieves a CER of 0.4472. Weighted sampling places higher importance on less common letters such as `x' and `q' and increases the average class accuracy by over 3\%. Additional gains are obtained through using data augmentation and per-frame loss, which further reduce the CER by 0.0043 and 0.0051, respectively. Incorporating \dataset{} yields the largest CER loss of 0.1899 as the number of training entries have increased from around 2k to 25k in total. Data augmentation increases data variability and is useful when there is limited data. Providing the model with per-frame letters enforces the correct letter at each time step. For example, when the letters `a' or `e' are equally likely due to similarity, the per-frame loss penalises the incorrect vowel. These results validate the importance of our additional annotations as well as the effectiveness of our model design choices. We further examine the input modalities. Combining lip and hand features achieves a CER of 0.2497, compared with 0.2945 using hands alone, demonstrating the benefit of lip features. Detailed results are provided in the supplementary material.

\section{Conclusion}
In this paper, we present a large-scale, fine-grained fingerspelling dataset for British Sign Language (BSL), constructed using an iterative fingerspelling annotation framework that leverages hand and lip features. Our dataset, \dataset{}, consists of 23,073 instances with continuous letter-level annotations. The proposed method precisely localizes fingerspelling sequences and identifies potentially incorrect annotations during the annotation process. Moreover, by incorporating robust hand keypoints, explicit modeling of bi-manual relationships, and lip features, our approach outperforms Transpeller by 0.331 and 0.144 on the CSLR and Transpeller testing datasets respectively in terms of character error rate (CER) on BSL fingerspelling recognition. We believe that our dataset and experimental results will facilitate further research on BSL, which poses unique challenges compared to other sign languages.

The limitation is that the current approach does not incorporate linguistic context during fingerspelling recognition. Integrating a letter-level language model could help resolve ambiguous predictions by correcting low-probability character sequences based on linguistic priors.
An interesting direction for future work is the development of language-agnostic fingerspelling models. By learning shared representations across multiple sign languages, such models could leverage larger and more diverse datasets, helping to mitigate data imbalance across languages and enabling support for under-resourced sign languages.

\paragraph{Acknowledgments.}
The BOBSL images in this paper are used with the kind permission of the BBC.
This work was supported by the UKRI EPSRC Programme Grant SignGPT EP/Z535370/1, a Royal Society
Research Professorship RSRP$\backslash$R$\backslash$241003 and an SNSF Postdoc.\ Mobility Fellowship P500PT\_225450.

\clearpage

\bibliography{custom}

@inproceedings{low_hands-_2025,
	title = {Hands-On: Segmenting Individual Signs from Continuous Sequences},
	url = {http://arxiv.org/abs/2504.08593},
	doi = {10.1109/FG61629.2025.11099255},
	shorttitle = {Hands-On},
	pages = {1--5},
	booktitle = {2025 {IEEE} 19th International Conference on Automatic Face and Gesture Recognition ({FG})},
	author = {Low, {JianHe} and Walsh, Harry and Sincan, Ozge Mercanoglu and Bowden, Richard},
	urldate = {2026-02-27},
	date = {2025-05-26},
	eprinttype = {arxiv},
	eprint = {2504.08593 [cs]},
}

@inproceedings{papadimitriou_seeing_2025,
	title = {Seeing in 2D, Thinking in 3D: 3D Hand Mesh-Guided Feature Learning for Continuous Fingerspelling},
	url = {https://openaccess.thecvf.com/content/ICCV2025W/CV4A11y/html/Papadimitriou_Seeing_in_2D_Thinking_in_3D_3D_Hand_Mesh-Guided_Feature_ICCVW_2025_paper.html},
	shorttitle = {Seeing in 2D, Thinking in 3D},
	eventtitle = {Proceedings of the {IEEE}/{CVF} International Conference on Computer Vision},
	pages = {6676--6685},
	author = {Papadimitriou, Katerina and Filntisis, Panagiotis and Retsinas, George and Potamianos, Gerasimos and Maragos, Petros},
	urldate = {2025-12-25},
	date = {2025},
	langid = {english},
}

@inproceedings{wheatland_analysis_2016,
    title = {Analysis in support of realistic timing in animated fingerspelling},
    url = {https://ieeexplore.ieee.org/document/7504777/},
    doi = {10.1109/VR.2016.7504777},
    urldate = {2025-04-23},
    booktitle = {2016 {IEEE} {Virtual} {Reality} ({VR})},
    author = {Wheatland, Nkenge and Abdullah, Ahsan and Neff, Michael and Jörg, Sophie and Zordan, Victor},
    month = mar,
    year = {2016},
    note = {ISSN: 2375-5334},
    pages = {309--310},
}

@misc{raude_tale_2024,
    title = {A {Tale} of {Two} {Languages}: {Large}-{Vocabulary} {Continuous} {Sign} {Language} {Recognition} from {Spoken} {Language} {Supervision}},
    shorttitle = {A {Tale} of {Two} {Languages}},
    url = {http://arxiv.org/abs/2405.10266},
    doi = {10.48550/arXiv.2405.10266},
    urldate = {2025-04-26},
    publisher = {arXiv},
    author = {Raude, Charles and Prajwal, K. R. and Momeni, Liliane and Bull, Hannah and Albanie, Samuel and Zisserman, Andrew and Varol, Gül},
    month = may,
    year = {2024},
    note = {arXiv:2405.10266 [cs]},
}

@incollection{kisacanin_recognition_2005,
    address = {New York},
    title = {Recognition of {Isolated} {Fingerspelling} {Gestures} {Using} {Depth} {Edges}},
    isbn = {978-0-387-27697-7},
    url = {http://link.springer.com/10.1007/0-387-27890-7_3},
    language = {en},
    urldate = {2025-04-25},
    booktitle = {Real-{Time} {Vision} for {Human}-{Computer} {Interaction}},
    publisher = {Springer-Verlag},
    author = {Feris, Rogerio and Turk, Matthew and Raskar, Ramesh and Tan, Kar-Han and Ohashi, Gosuke},
    editor = {Kisačanin, Branislav and Pavlović, Vladimir and Huang, Thomas S.},
    year = {2005},
    doi = {10.1007/0-387-27890-7_3},
    pages = {43--56},
}

@inproceedings{liwicki_automatic_2009,
    title = {Automatic recognition of fingerspelled words in {British} {Sign} {Language}},
    url = {https://ieeexplore.ieee.org/document/5204291/},
    doi = {10.1109/CVPRW.2009.5204291},
    urldate = {2025-04-26},
    booktitle = {2009 {IEEE} {Computer} {Society} {Conference} on {Computer} {Vision} and {Pattern} {Recognition} {Workshops}},
    author = {Liwicki, Stephan and Everingham, Mark},
    month = jun,
    year = {2009},
    note = {ISSN: 2160-7516},
    pages = {50--57},
}

@article{kumar_deaf-bsl_2022,
    title = {{DEAF}-{BSL}: {Deep} {lEArning} {Framework} for {British} {Sign} {Language} recognition},
    volume = {21},
    issn = {2375-4699, 2375-4702},
    shorttitle = {{DEAF}-{BSL}},
    url = {https://dl.acm.org/doi/10.1145/3513004},
    doi = {10.1145/3513004},
    language = {en},
    number = {5},
    urldate = {2025-04-29},
    journal = {ACM Transactions on Asian and Low-Resource Language Information Processing},
    author = {Kumar, Krishan},
    month = sep,
    year = {2022},
    pages = {1--14},
}

@inproceedings{goh_dynamic_2006,
    title = {Dynamic {Fingerspelling} {Recognition} using {Geometric} and {Motion} {Features}},
    url = {https://ieeexplore.ieee.org/document/4107136/},
    doi = {10.1109/ICIP.2006.313114},
    urldate = {2025-04-29},
    booktitle = {2006 {International} {Conference} on {Image} {Processing}},
    author = {Goh, Paul and Holden, Eun-jung},
    month = oct,
    year = {2006},
    note = {ISSN: 2381-8549},
    pages = {2741--2744},
}

@inproceedings{pinnington_machine_2024,
    title = {Machine {Learning} in {ASL} {Fingerspelling} {Recognition}: {A} {Literature} {Review}},
    shorttitle = {Machine {Learning} in {ASL} {Fingerspelling} {Recognition}},
    url = {https://ieeexplore.ieee.org/document/10830860/},
    doi = {10.1109/CINTI63048.2024.10830860},
    urldate = {2025-04-29},
    booktitle = {2024 {IEEE} 24th {International} {Symposium} on {Computational} {Intelligence} and {Informatics} ({CINTI})},
    author = {Pinnington, Jamie and Souag, Amina and Hannan Bin Azhar, M A},
    month = nov,
    year = {2024},
    note = {ISSN: 2471-9269},
    pages = {000055--000062},
}

@article{graves_connectionist_nodate,
    title = {Connectionist {Temporal} {Classiﬁcation}: {Labelling} {Unsegmented} {Sequence} {Data} with {Recurrent} {Neural} {Networks}},
    language = {en},
    author = {Graves, Alex and Fernandez, Santiago and Gomez, Faustino and Schmidhuber, Jurgen},
}

@misc{pavlakos_reconstructing_2023,
    title = {Reconstructing {Hands} in {3D} with {Transformers}},
    url = {http://arxiv.org/abs/2312.05251},
    doi = {10.48550/arXiv.2312.05251},
    urldate = {2025-04-29},
    publisher = {arXiv},
    author = {Pavlakos, Georgios and Shan, Dandan and Radosavovic, Ilija and Kanazawa, Angjoo and Fouhey, David and Malik, Jitendra},
    month = dec,
    year = {2023},
    note = {arXiv:2312.05251 [cs]},
}

@inproceedings{danko_recognition_2019,
    title = {Recognition of the {Hungarian} fingerspelling alphabet using {Recurrent} {Neural} {Network}},
    url = {https://ieeexplore.ieee.org/document/8782725/},
    doi = {10.1109/SAMI.2019.8782725},
    urldate = {2025-04-29},
    booktitle = {2019 {IEEE} 17th {World} {Symposium} on {Applied} {Machine} {Intelligence} and {Informatics} ({SAMI})},
    author = {Dankó, Bence and Kertész, Gábor},
    month = jan,
    year = {2019},
    pages = {251--256},
}

@article{oz_american_2011,
    series = {Infrastructures and {Tools} for {Multiagent} {Systems}},
    title = {American {Sign} {Language} word recognition with a sensory glove using artificial neural networks},
    volume = {24},
    issn = {0952-1976},
    url = {https://www.sciencedirect.com/science/article/pii/S0952197611001230},
    doi = {10.1016/j.engappai.2011.06.015},
    number = {7},
    urldate = {2025-04-29},
    journal = {Engineering Applications of Artificial Intelligence},
    author = {Oz, Cemil and Leu, Ming C.},
    month = oct,
    year = {2011},
    pages = {1204--1213},
}

@inproceedings{ricco_fingerspelling_2010,
    address = {Berlin, Heidelberg},
    title = {Fingerspelling {Recognition} through {Classification} of {Letter}-to-{Letter} {Transitions}},
    isbn = {978-3-642-12297-2},
    doi = {10.1007/978-3-642-12297-2_21},
    language = {en},
    booktitle = {Computer {Vision} – {ACCV} 2009},
    publisher = {Springer},
    author = {Ricco, Susanna and Tomasi, Carlo},
    editor = {Zha, Hongbin and Taniguchi, Rin-ichiro and Maybank, Stephen},
    year = {2010},
    pages = {214--225},
}

@article{sutton-spence_variation_1990,
    title = {Variation and recent change in fingerspelling in {British} {Sign} {Language}},
    volume = {2},
    copyright = {https://www.cambridge.org/core/terms},
    issn = {0954-3945, 1469-8021},
    url = {https://www.cambridge.org/core/product/identifier/S0954394500000399/type/journal_article},
    doi = {10.1017/S0954394500000399},
    language = {en},
    number = {3},
    urldate = {2025-04-29},
    journal = {Language Variation and Change},
    author = {Sutton-Spence, Rachel and Woll, Bencie and Allsop, Lorna},
    month = oct,
    year = {1990},
    pages = {313--330},
}

@article{brennan_british_1979,
    title = {A {British} {Sign} {Language} {Research} {Project}},
    issn = {0302-1475},
    url = {https://www.jstor.org/stable/26203628},
    number = {24},
    urldate = {2025-04-30},
    journal = {Sign Language Studies},
    author = {Brennan, Mary and Colville, Martin},
    year = {1979},
    note = {Publisher: Gallaudet University Press},
    pages = {253--272},
}

@article{rambhau_recognition_2013,
    title = {Recognition of {Two} {Hand} {Gestures} of word in {British} {Sign} {Language} ({BSL})},
    volume = {3},
    language = {en},
    number = {10},
    author = {Rambhau, Pingale Prerna},
    year = {2013},
}

@inproceedings{papadimitriou_multimodal_2024,
    title = {Multimodal {Continuous} {Fingerspelling} {Recognition} via {Visual} {Alignment} {Learning}},
    url = {https://www.isca-archive.org/interspeech_2024/papadimitriou24_interspeech.html},
    doi = {10.21437/Interspeech.2024-1966},
    language = {en},
    urldate = {2025-05-02},
    booktitle = {Interspeech 2024},
    publisher = {ISCA},
    author = {Papadimitriou, Katerina and Potamianos, Gerasimos},
    month = sep,
    year = {2024},
    pages = {922--926},
}

@inproceedings{fayyazsanavi_fingerspelling_2024,
    address = {Waikoloa, HI, USA},
    title = {Fingerspelling {PoseNet}: {Enhancing} {Fingerspelling} {Translation} with {Pose}-{Based} {Transformer} {Models}},
    copyright = {https://doi.org/10.15223/policy-029},
    isbn = {9798350370287},
    shorttitle = {Fingerspelling {PoseNet}},
    url = {https://ieeexplore.ieee.org/document/10495714/},
    doi = {10.1109/WACVW60836.2024.00123},
    language = {en},
    urldate = {2025-05-02},
    booktitle = {2024 {IEEE}/{CVF} {Winter} {Conference} on {Applications} of {Computer} {Vision} {Workshops} ({WACVW})},
    publisher = {IEEE},
    author = {Fayyazsanavi, Pooya and Nejatishahidin, Negar and Košecká, Jana},
    month = jan,
    year = {2024},
    pages = {1120--1130},
}

@misc{prajwal_prajwalkrtranspeller_2025,
    title = {prajwalkr/transpeller},
    url = {https://github.com/prajwalkr/transpeller},
    urldate = {2025-05-13},
    author = {Prajwal},
    month = jan,
    year = {2025},
    note = {original-date: 2023-06-12T19:06:33Z},
}

@InProceedings{prajwal2022weakly,
  author       = "K R Prajwal and Hannah Bull and Liliane Momeni and Samuel Albanie and G{\"u}l Varol and Andrew Zisserman",
  title        = "Weakly-supervised Fingerspelling Recognition in British Sign Language Videos",
  booktitle    = "British Machine Vision Conference",
  year         = "2022",
}

@inproceedings{albanie2020bsl,
  title={BSL-1K: Scaling up co-articulated sign language recognition using mouthing cues},
  author={Albanie, Samuel and Varol, G{\"u}l and Momeni, Liliane and Afouras, Triantafyllos and Chung, Joon Son and Fox, Neil and Zisserman, Andrew},
  booktitle={European conference on computer vision},
  pages={35--53},
  year={2020},
  organization={Springer}
}

@article{albanie2021bbc,
  title={{BBC-Oxford} british sign language dataset},
  author={Albanie, Samuel and Varol, G{\"u}l and Momeni, Liliane and Bull, Hannah and Afouras, Triantafyllos and Chowdhury, Himel and Fox, Neil and Woll, Bencie and Cooper, Rob and McParland, Andrew and others},
  journal={arXiv preprint arXiv:2111.03635},
  year={2021}
}

@article{adam2014method,
  title={A method for stochastic optimization},
  author=  {Kingma, Diederik P. and Ba, Jimmy},
  journal={arXiv preprint arXiv:1412.6980},
  volume={1412},
  number={6},
  year={2014}
}

@inproceedings{kapitanov2023slovo,
  title={Slovo: Russian sign language dataset},
  author={Kapitanov, Alexander and Karina, Kvanchiani and Nagaev, Alexander and Elizaveta, Petrova},
  booktitle={International Conference on Computer Vision Systems},
  pages={63--73},
  year={2023},
  organization={Springer}
}

@inproceedings{georg2025fsboard,
  title={FSboard: Over 3 million characters of ASL fingerspelling collected via smartphones},
  author={Georg, Manfred and Tanzer, Garrett and Uboweja, Esha and Hassan, Saad and Shengelia, Maximus and Sepah, Sam and Forbes, Sean and Starner, Thad},
  booktitle={Proceedings of the Computer Vision and Pattern Recognition Conference},
  pages={13897--13906},
  year={2025}
}

@article{shen2023auslan,
  title={Auslan-daily: Australian sign language translation for daily communication and news},
  author={Shen, Xin and Yuan, Shaozu and Sheng, Hongwei and Du, Heming and Yu, Xin},
  journal={Advances in Neural Information Processing Systems},
  volume={36},
  pages={80455--80469},
  year={2023}
}

@article{lata2024one,
  title={One-Stage-TFS: Thai One-Stage Fingerspelling Dataset for Fingerspelling Recognition Frameworks},
  author={Lata, Siriwiwat and Phiphitphatphaisit, Sirawan and Okafor, Emmanuel and Surinta, Olarik},
  journal={arXiv preprint arXiv:2411.02768},
  year={2024}
}

@inproceedings{shi2018american,
  title={American sign language fingerspelling recognition in the wild},
  author={Shi, Bowen and Del Rio, Aurora Martinez and Keane, Jonathan and Michaux, Jonathan and Brentari, Diane and Shakhnarovich, Greg and Livescu, Karen},
  booktitle={2018 IEEE Spoken Language Technology Workshop (SLT)},
  pages={145--152},
  year={2018},
  organization={IEEE}
}

@inproceedings{shi2019fingerspelling,
  title={Fingerspelling recognition in the wild with iterative visual attention},
  author={Shi, Bowen and Rio, Aurora Martinez Del and Keane, Jonathan and Brentari, Diane and Shakhnarovich, Greg and Livescu, Karen},
  booktitle={Proceedings of the IEEE/CVF International Conference on Computer Vision},
  pages={5400--5409},
  year={2019}
}

@inproceedings{ma2023auto,
  title={{Auto-AVSR:} Audio-visual speech recognition with automatic labels},
  author={Ma, Pingchuan and Haliassos, Alexandros and Fernandez-Lopez, Adriana and Chen, Honglie and Petridis, Stavros and Pantic, Maja},
  booktitle={ICASSP 2023-2023 IEEE International Conference on Acoustics, Speech and Signal Processing (ICASSP)},
  pages={1--5},
  year={2023},
  organization={IEEE}
}
\clearpage

\appendix

\setcounter{figure}{0}
\setcounter{table}{0}

\renewcommand{\thefigure}{A\arabic{figure}}
\renewcommand{\thetable}{A\arabic{table}}

\section{Background on BSL Fingerspelling}

British Sign Language (BSL) uses a unique bi-manual fingerspelling system, illustrated in Figure 2. All letters except ‘c’ require the use of both hands~\cite{kumar_deaf-bsl_2022}. Additionally, all but the letters 'h' and 'j' are represented using static signs, where the letter is conveyed through a specific hand shape and contact point between the two hands rather than motion. 

In practice, the exact hand poses are not always identical to those shown in Figure~2. For example, the letter ‘a’ is formed by contact between the right-hand index finger and the left-hand thumb. While the left hand is often depicted as outstretched, its precise configuration is not semantically meaningful and may vary across individuals.
The same letter can be produced in multiple ways depending on the signer~\cite{sutton-spence_variation_1990}, as illustrated in Figure~3. This flexibility allows for natural variation in signing style but also complicates recognition, as small changes in finger angle, hand orientation, or contact location can lead to ambiguous or misclassified signs. In addition, several letters, such as ‘l’, ‘m’, ‘n’, ‘r’, and ‘v’, are particularly difficult to 
distinguish~\cite{liwicki_automatic_2009}, as they differ only in the number and configuration of fingers resting against the palm. This challenge is further exacerbated by frequent finger occlusions.

\section{Annotation Method Details}

In order to train a frame-wise classifier, we assign a letter to each frame. We employ three methods to undertake this task, which is illustrated in Figure~\ref{fig:lettercomp}.

\paragraph{Equal number of frames per letter} 

Word-level annotations do not specify frame-to-letter correspondences. To address this, we adopt a simple initialization strategy that evenly divides the frames of each fingerspelling sequence among its constituent letters. For example, if the word “Darwin” is fingerspelt over 60 frames, each letter (‘d’, ‘a’, ‘r’, ‘w’, ‘i’, ‘n’) is assigned 10 frames.

\paragraph{Transitions between letters} 

In the CSLR dataset \cite{raude_tale_2024}, the exact letter sequence for each fingerspelling instance is known. Leveraging this information, we use our classifier to detect transition points between successive letters. Specifically, given the expected next letter, we identify the frame at which the predicted probability of the next letter exceeds that of the current letter. This enables segmentation based on model confidence rather than fixed-length temporal divisions. However, this approach restricts the usable data to samples with complete letter-level annotations.

\paragraph{Maximum probability} 

An alternative approach assigns a letter label to each frame based on maximum prediction confidence, and can be applied across the entire dataset (where a word is associated with each interval). Specifically, the classifier’s softmax layer outputs a probability distribution over letters for each frame. We first retain predictions whose maximum probability exceeds a threshold of 0.35, and then assign the letter with the highest probability among these candidates.
As illustrated in Figure~\ref{fig:lettercomp}, when the predicted probability for the letter ‘l’ falls below the 0.35 threshold, we retain the previously assigned letter. The threshold value of 0.35 is chosen heuristically through visual inspection, selecting the value that yields the most accurate and temporally consistent predictions.

\begin{figure}[t]
    \centering
    \includegraphics[width=0.3\linewidth]{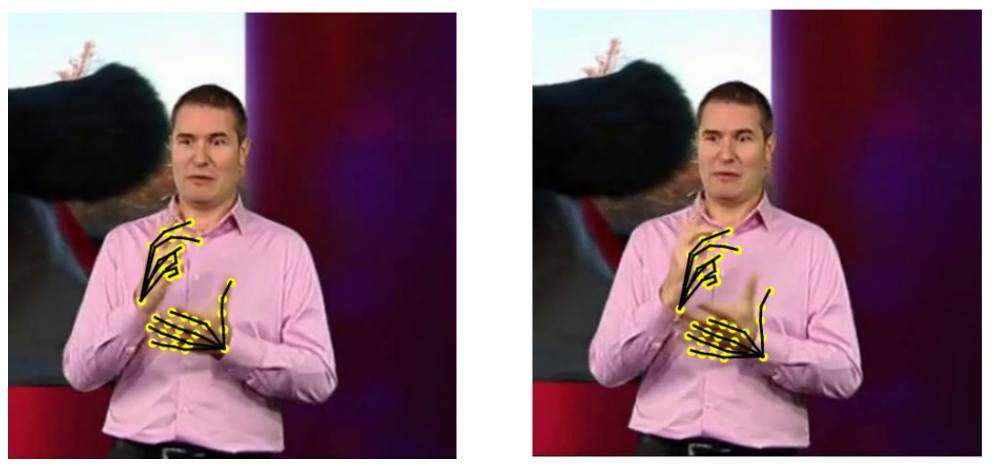}
    \vspace{0.02\textwidth}
    \caption{\small Data Augmentation Examples. 2D keypoints with no augmentation (left) and 12 pixels (right) in horizontal and vertical directions.}
    \label{fig:augmentation}
\end{figure}

\begin{figure*}
    \centering
    \includegraphics[width=0.8\linewidth]{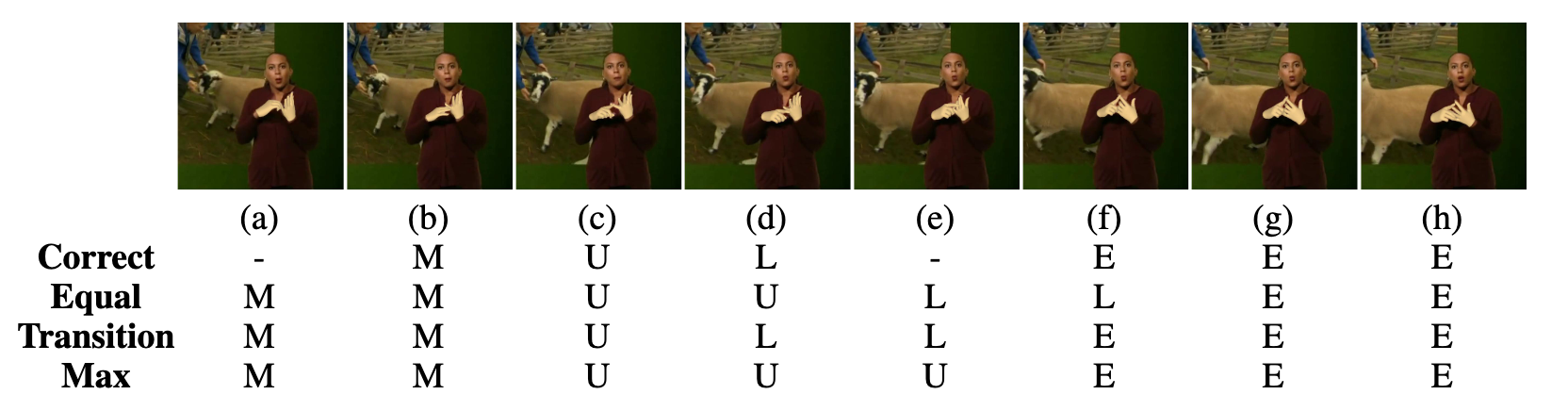}
    \vspace{6pt}
    \caption{\small Visual representation of each method used to annotate letters in CSLR dataset. The word `mule' is being spelt. The `correct' annotation shows the actual letters being spelt in the frame with transition periods shown with a dashed line.}
    \label{fig:lettercomp}
\end{figure*}

\begin{figure*}
    \centering
    \includegraphics[width=0.8\linewidth]{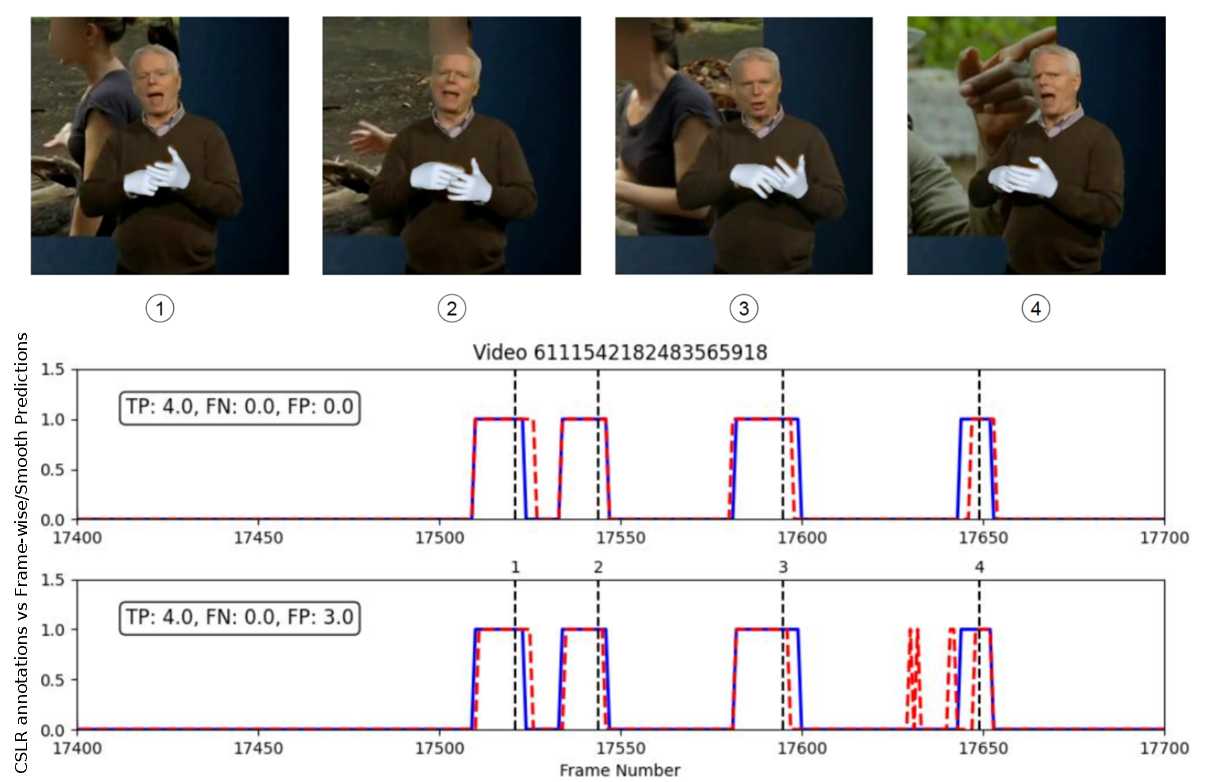}
    \vspace{5pt}
    \caption{\small Example visualisation on the CSLR test set. The dashed lines correspond to the frames displayed at the top of the figure. The blue lines show the ground truth. There are four true positive events with \(IoU \geq 0.5\). The bottom panel depicts the frame-wise predictions and the top panel shows the results after applying the sliding window, both in red.}
    \label{truepos}
\end{figure*}

\begin{figure}
    \centering
    \small
    \includegraphics[width=0.8\linewidth]{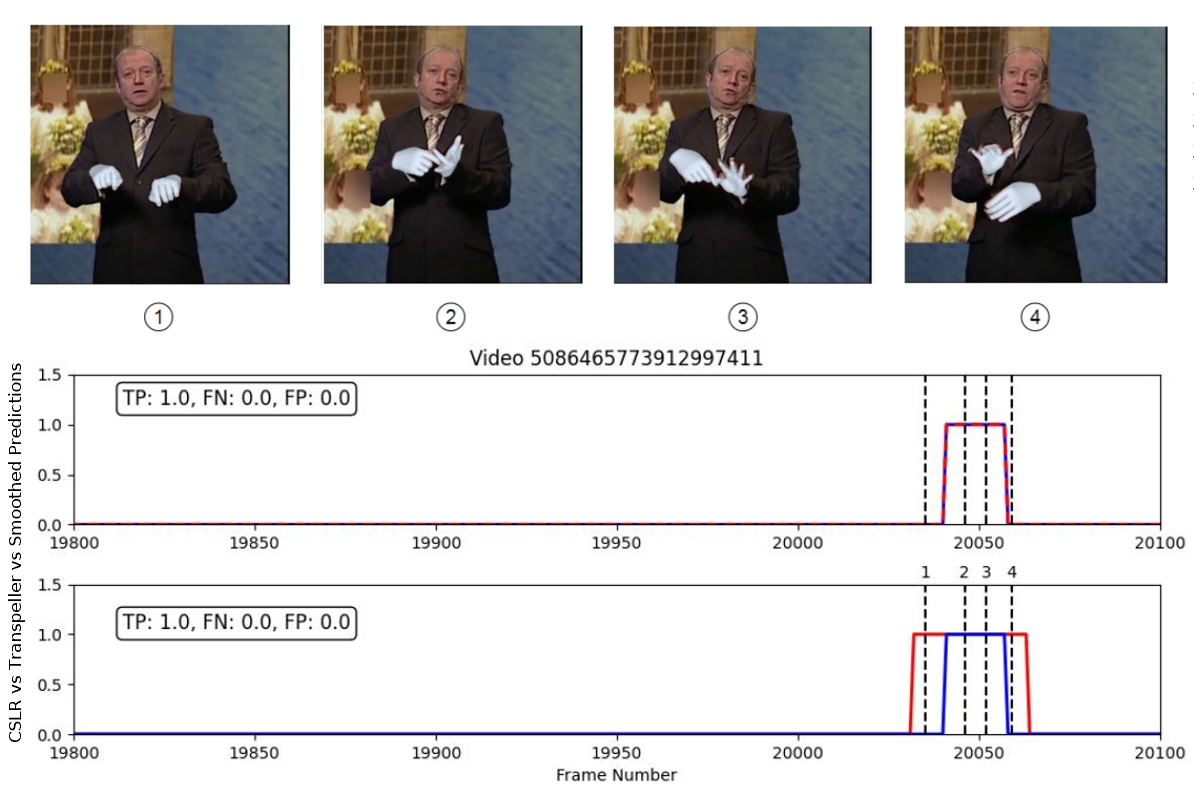}
    \vspace{5pt}
    \caption{\small Visualisation of an example with both CSLR and Transpeller annotations. The dashed lines correspond to the frames displayed at the top of the figure. The top panel shows the ground truth from CSLR in blue and our smoothed predictions in a red dashed line. The bottom panel shows the Transpeller automatic annotations in solid red and the ground truth from CSLR in blue. We observe that the Transpeller annotations span much wider intervals that include non-fingerspelling frames, whereas our predictions closely match the ground truth and precisely localise the frames where fingerspelling occurs.}
    \label{tony}
\end{figure}

\section{Implementation Details}

\begin{table}[t]
  \caption{\small Statistics of the automatically annotated Transpeller dataset after successive filtering steps.} 
  \vspace{6pt}

  \centering
  \small
  \begin{tabularx}{0.8\columnwidth}{X r}
    \toprule
    \textbf{Subset} & \textbf{\# Entries} \\
    \midrule
    (1) Full noisy labels & 157{,}182 \\
    (2) Removing non-existing videos dataset & 154{,}781\\
    (3) Removing entries that have no fingerspelling or multiple fingerspelling events & \textbf{133{,}909}\\
    (3) Keeping entries from the training videos  & 113{,}100 \\
    (4) Removing missing word annotations & 58{,}335 \\
    (5) Restricting entries with less than 0.3 CER between the associated and predicted word & \textbf{23{,}073} \\
    \bottomrule
  \end{tabularx}
  \label{tab:transpellerdata}
\end{table}

\subsection{Subset Selection}

Out of the 157,182 Transpeller entries, we retain 23,073 valid instances to construct high-quality fingerspelling annotations, as summarized in Table~\ref{tab:transpellerdata}. Specifically, we first remove 2,401 entries that do not exist in the BOBSL dataset~\cite{albanie2020bsl} version 4. We use our detector to remove entries that lack fingerspelling or have multiple events in one interval. This forms our fingerspelling temporal boundaries dataset. We exclude 19,097 entries belonging to the BOBSL testing videos to avoid data leakage. 
Next, we discard 54,765 entries that lack annotations. 
Finally, after the iterative annotation process, we accept 23,073 samples, which guarantee high-quality fingerspelling annotations both in temporal granularity and letter labels. 

\begin{table}[t]
\centering
\small

\caption{\small Feature ablation study showing the improvement in CER when using combined features. Used our \dataset{}, per-frame loss and data augmentation.}
\vspace{0.05\textwidth}

\begin{tabular}{|c|c|c|c|}
\hline
\textbf{Only Lip features} &  \textbf{Only Hand features} & \textbf{Combined features} & \textbf{CER$\downarrow$} \\
\hline

\cmark & \xmark & \xmark & 0.8756 \\
\xmark & \cmark & \xmark & 0.2945 \\
\xmark & \xmark & \cmark & 0.2497 \\
\hline
\end{tabular}

\label{tab:ablationfeatures}
\end{table}

\begin{figure}[h!]
    \centering
    \includegraphics[width=0.5\linewidth]{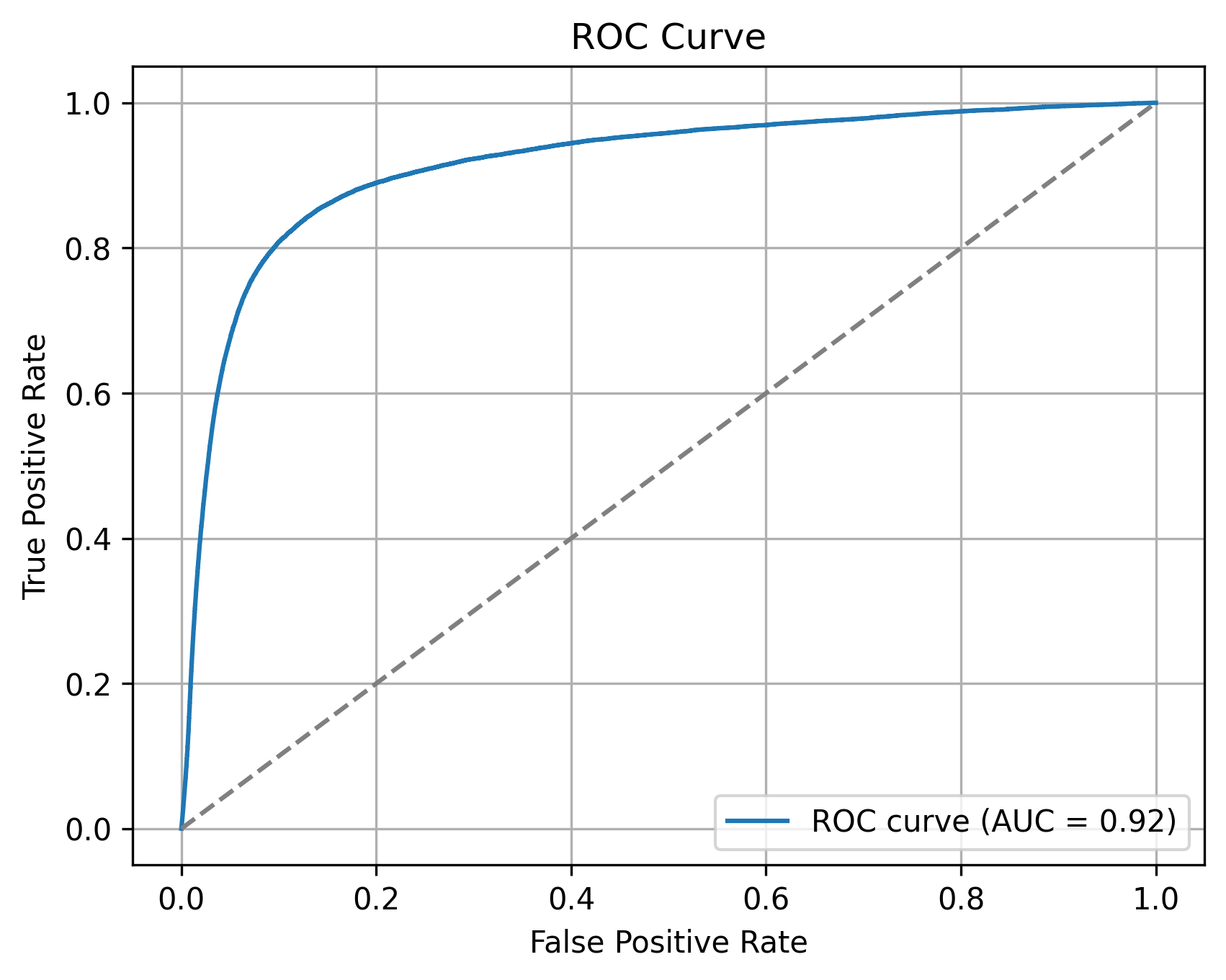}
    \caption{Fingerspelling detector ROC curve. The detector achieves an area under the curve (AUC) of 0.92. }
    \label{fig:roc}
\end{figure}

\section{Detection Illustrations}

In this section, we present the results of our fingerspelling detector. For more examples, please see the supplementary video. The detector achieves an average class accuracy of 87.94\% on the CSLR test set, as shown by the ROC curve in Figure~\ref{fig:roc}.
We apply a sliding-window smoothing strategy to the frame-wise predictions. As illustrated in Figure~\ref{truepos}, the classifier accurately identifies the start and end frames of fingerspelling segments. The benefit of the sliding window is evident in the removal of spurious predictions around frame 17640, reducing the number of false positives by three. The effect of our fingerspelling detector is shown in Figure~\ref{tony}, the fingerspelling interval is shortened by around 20 frames or 0.8 seconds. 

Importantly, the model is also able to detect valid fingerspelling segments that are absent from the original CSLR annotations~\cite{prajwal2022weakly}. This demonstrates the potential of the model not only as a classifier but also as a tool for identifying missed or inconsistent annotations.

Most false positives are short, typically lasting one to two seconds. These brief detections often correspond to lexical signs or gestures that visually resemble fingerspelling, such as moments when the hands are close together or adopt letter-like shapes (e.g., curved or outstretched configurations). In contrast, longer detected fingerspelling intervals are more frequently attributable to annotation errors, suggesting that fingerspelling may indeed have been present but was not labelled.

\end{document}